\documentclass{article}

\PassOptionsToPackage{numbers,compress}{natbib}

\usepackage[preprint]{neurips_2026}

\usepackage[utf8]{inputenc} 
\usepackage[T1]{fontenc}    
\usepackage{hyperref}       
\usepackage{url}            
\usepackage{booktabs}       
\usepackage{amsfonts,amsmath}       
\usepackage{nicefrac}       
\usepackage{microtype}      
\usepackage{xcolor}         
\usepackage{algorithm}
\usepackage{algpseudocode}
\usepackage{enumitem}
\usepackage{graphicx}

\DeclareRobustCommand{\citet}[1]{\citeauthor{#1}, \citeyear{#1}~\citep{#1}}

\title{Trust but Verify: Prover-Verifier Deliberation for Selective LLM Prediction}

\author{%
  Jo\~ao Sedoc \\
  New York University \\
  \texttt{jsedoc@stern.nyu.edu} \\
  \And
 Baotong Zhang \\
  New York University \\
  \texttt{bz2274@stern.nyu.edu} \\
  \AND
  Dean Foster \\
  \texttt{dean@foster.net} \\
}

\begin{document}

\maketitle

\begin{abstract}
Reliably knowing when a language model is correct is almost as important as being correct. We introduce prover-verifier deliberation (PVD), an inference-time protocol grounded in interactive proof theory, as a mechanism for selective prediction: the protocol produces both an answer and a structured confidence verdict, allowing a system to report high-confidence answers while abstaining on uncertain cases. In each dialogue, a prover defends a candidate answer through checkable sub-claims while a verifier issues targeted challenges and returns \textsc{Accept}, \textsc{Challenge}, or \textsc{Reject}. Because frozen language models are imperfect provers and verifiers operating over a noisy channel, formal soundness and completeness guarantees do not transfer; instead, we characterize the protocol empirically through its coverage-precision behavior. Our main experiment uses Claude Sonnet 4.6 as prover and Claude Haiku 4.5 as verifier on GPQA Diamond. Questions accepted with no answer revision, which we call Accept + No Change (ANC), are reported as the high-confidence subset; we evaluate this subset by its precision and coverage. PVD achieves HC-Prec 84.2\% at HC-Cov 77\% with roughly three LLM calls per question. ANC separates reliable from unreliable answers, yielding a $\sim$30pp HC-Prec gap over the non-ANC complement. Robustness experiments with GPT and Gemini pairings show that high HC-Prec can transfer across model families, while verifier strictness and domain competence largely determine the size of the selection gap. On Humanity's Last Exam, weaker prover-verifier pairings can collapse or invert the ANC signal, illustrating a practical failure mode when the verifier operates outside its effective region. Comparisons with self-consistency, universal self-consistency, multi-agent debate, and Reflexion suggest that prover-verifier deliberation supplies a distinct argument-defensibility signal for selective prediction rather than merely reproducing sample agreement or debate consensus.\footnote{Code for reproduction and Claude Code skill available at \url{https://github.com/jsedoc/prover-verifier-deliberation}.}
\end{abstract}

\section{Introduction}
\label{sec:intro}

Reliably knowing when a language model is correct is often as important as being correct. A system that answers 90\% of questions accurately but cannot distinguish its correct answers from its errors is far less useful in deployment than one that answers 60\% of questions with near-certainty and abstains otherwise. This observation motivates selective prediction \citep{geifman2017selective}: rather than producing a single accuracy figure, a well-calibrated system should trade coverage for precision, reporting answers only where it is confident. Despite its practical importance, selective prediction for LLMs remains underexplored relative to the large literature on improving raw accuracy through inference-time compute.

Inference-time reasoning methods have advanced rapidly since chain-of-thought prompting \citep{wei2022chain} demonstrated that intermediate reasoning steps improve accuracy on complex tasks. Subsequent work spans self-consistency \citep{wang2023selfconsistency}, which aggregates multiple independent samples via majority vote; Reflexion \citep{shinn2023reflexion}, which refines answers through verbal self-reflection; Universal Self-Consistency \citep{chen2024usc}, which uses an LLM aggregator to select the most coherent candidate; and multi-agent debate \citep{du2024debate}, where multiple model instances argue toward consensus. These methods improve accuracy, but their confidence signals are weak: self-consistency's agreement rate and debate's consensus are at best indirect proxies for correctness, and none was designed to produce a principled abstain/report decision.

A more structured approach comes from interactive proof theory \citep{Goldwasser1989TheKC}. In the interactive proof (IP) / zero-knowledge proof (ZKP) paradigm, a computationally powerful \textbf{prover} convinces a skeptical \textbf{verifier} of a claim's truth through a structured challenge-response dialogue; the verifier's final verdict carries formal soundness and completeness guarantees. Recent work has explored this paradigm in learning systems: \citet{hammond2025neural} study Neural Interactive Proofs (NIP), proving that \textsf{NIP} $=$ \textsf{PSPACE}, and training prover-verifier pairs via approximate Stackelberg optimization on verifiable tasks. \citet{zhang2025human} examine human-LLM interactive proof settings. These works focus on trained agents and formal verifiability. Our work takes a complementary direction: we ask whether the \textbf{structure} of interactive proof dialogue (without training, without verifiable ground truth, and over open-domain knowledge questions) can serve as a practical calibration mechanism for frozen LLMs.

We introduce Prover-Verifier Deliberation (PVD), a challenge-guided selective-prediction protocol in which a prover defends a candidate answer and a verifier issues targeted challenges before returning an Accept, Challenge, or Reject verdict. The dialogue terminates when the verifier reaches a verdict or a fatigue limit is hit. Because language models are imperfect agents operating over a noisy channel rather than computationally bounded reasoners, hard soundness and completeness guarantees do not transfer; instead, we characterize the protocol empirically as a selective predictor. We introduce Accept + No Change (ANC) — questions where the prover's answer is accepted without revision — as a coverage-precision operating point, and evaluate it against standard baselines across two benchmarks and multiple hosted model pairings.

\begin{figure}[t]
\centering
\includegraphics[width=0.92\textwidth]{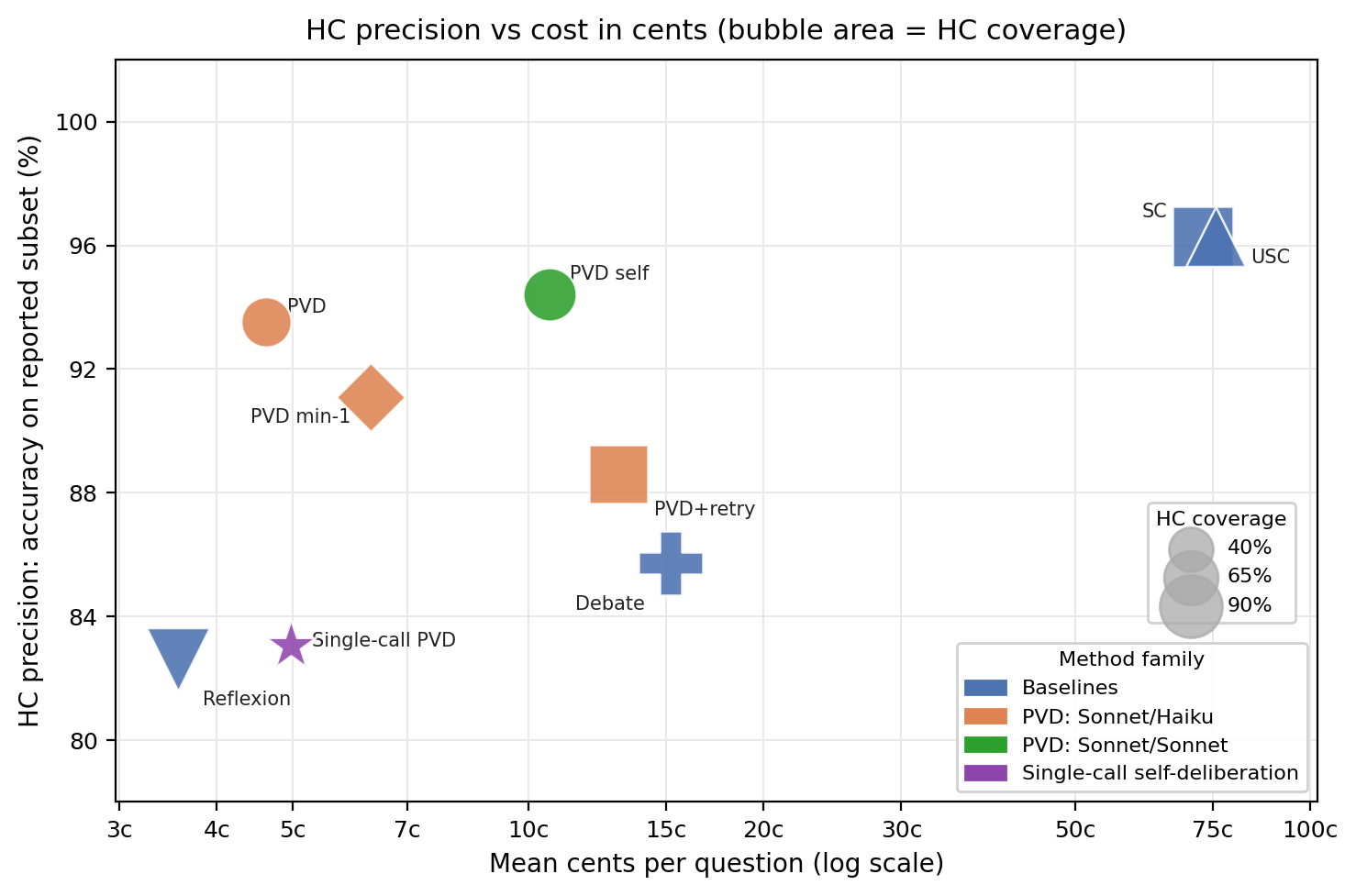}
\caption{Cost--precision tradeoff on GPQA Diamond for the clean RTTR runs. Each point is a method; the $y$-axis shows HC-Prec (accuracy on the reported high-confidence subset), the $x$-axis shows mean cents per question on a log scale, and bubble area encodes HC-Cov (fraction of questions reported). PVD variants occupy the high-precision region while exposing a controllable cost--coverage tradeoff: stricter or retry-based protocols spend more compute to report on more questions, while lower-cost variants abstain more aggressively.}
\label{fig:rttr-cost}
\end{figure}

Our main findings are: (i) On GPQA Diamond \citep{rein2024gpqa}, ANC questions achieve HC-Prec 84--98\% at HC-Cov 43--77\%, with a +6.6--34.8pp gap over the non-ANC complement. (ii) The strongest ANC signal comes from a within-family capability-asymmetric pairing: GPT-5.4 prover with GPT-5.4-mini verifier yields HC-Prec 97.6\% at HC-Cov 43\% (+34.8pp over the complement). (iii) On Humanity's Last Exam (HLE) \citep{phan2025lastexam}, weak prover-verifier pairings can collapse or invert the Accept/Reject accuracy gap, providing a natural diagnostic for when deliberation cannot be trusted. (iv) Compared to self-consistency ($k{=}8$), multi-agent debate, Universal Self-Consistency, and Reflexion, PVD occupies a distinct and favorable position on the accuracy–calibration–efficiency Pareto frontier (Figure~\ref{fig:rttr-cost}).

\section{Background \& Related Work}
\label{sec:rel-work}

Our work connects five lines of research: selective prediction and abstention, calibration and uncertainty estimation for language models \ref{sec:selective-prediction}, inference-time reasoning and aggregation \ref{sec:inference-time-reasoning}, verification and multi-agent deliberation \ref{sec:verification}, and interactive proofs and prover-verifier games. We position prover-verifier deliberation (PVD) as a selective-prediction  protocol: the goal is not only to improve the final answer, but to produce a structured report/abstain signal from the interaction itself. In this sense, PVD draws conceptual structure from interactive proof systems while remaining an empirical inference-time method for frozen, general-purpose LLMs.

\subsection{Interactive Proofs and Prover-Verifier Games}

Interactive proof systems formalize verification as a multi-round protocol between a powerful prover and a computationally bounded probabilistic verifier \citep{Goldwasser1989TheKC}. Classical results such as IP $=$ PSPACE show that interaction and randomness can greatly expand what can be efficiently verified \citep{shamir1992ip}. These results provide the conceptual template for PVD: a claim is not accepted merely because it is asserted, but because it survives a challenge-response process.

The closest conceptual comparison is Neural Interactive Proofs \citep{hammond2025neural}. That work studies neural prover-verifier protocols as learned interaction systems, develops a unifying theoretical framework, and evaluates trained protocols on tasks with verifiable structure. Human-LLM interactive proof settings provide another related perspective, focusing on the conditions under which a human verifier can adjudicate LLM-generated arguments \citep{zhang2025human}.

PVD occupies a complementary niche. We do not train provers or verifiers, assume a ground-truth oracle during interaction, or claim formal soundness and completeness. Frozen LLMs are imperfect provers and imperfect verifiers: they may hallucinate, miss valid challenges, reject correct arguments, or accept incorrect ones. Consequently, the formal guarantees of interactive proofs do not transfer. We instead ask an empirical selective-prediction question: \textbf{when the interactive-proof structure is instantiated with off-the-shelf LLMs on open-domain expert questions, does the verifier's verdict separate reliable answers from answers that should be withheld?}

\subsection{Soundness, Completeness, and the Effective Verifier}
\label{sec:completeness}

Interactive proof systems carry two formal guarantees. \textbf{Completeness} requires that an honest prover with genuine knowledge can always convince the verifier: for a true claim $s \in L$, the verifier accepts with probability at least $1 - p^T$ after $T$ rounds, where $p < 1$ is the per-round probability of challenge. \textbf{Soundness} requires that a mistaken or deceptive prover cannot fabricate conviction: for a false claim $s \notin L$, the probability of acceptance is bounded above by $q^T$, where $q < 1$ is the per-round slip-through probability. Both error rates decrease exponentially with interaction depth.

\citet{zhang2025human} formalize the conditions under which these guarantees hold when the verifier is human and the prover is an LLM. Their central requirement is the \textbf{Effective Verifier} assumption: the verifier must be able to confirm valid responses and identify errors within a finite fatigue limit $T$. Three conditions govern whether this assumption is satisfied in practice.

\textbf{Asymmetric cognitive load.} The protocol is feasible precisely because the prover bears the burden of full proof generation while the verifier only checks individual steps locally — confirming that each sub-claim is valid and follows from the prior exchange. This asymmetry keeps the verification task tractable even for resource-bounded agents: step-wise evaluation is far less demanding than global correctness judgment.

These conditions do not automatically hold when the verifier is a frozen LLM. An LLM verifier may issue spurious rejections (violating completeness) or fail to detect a flawed argument (violating soundness) whenever it operates near its own competence ceiling on the domain in question. This analysis motivates our empirical framing: rather than claiming formal soundness and completeness, we ask whether the interactive-proof structure, instantiated with frozen LLMs, produces verdicts that \emph{empirically} separate reliable answers from unreliable ones. The collapse of the \textsc{Accept}/\textsc{Reject} accuracy gap observed on Humanity's Last Exam is precisely the empirical signature of an empty effective region: the verifier lacks the domain knowledge $|I|$ to maintain $c_i$ at levels sufficient for either soundness or completeness, and its verdict degrades to near-chance. The strong GPQA results, by contrast, arise in Chemistry and Physics domains where the verifier's knowledge is sufficient to sustain adversarial challenge across many rounds.

\section{Prover-Verifier Deliberation}
\label{sec:protocol}

\subsection{Protocol}

Prover-Verifier Deliberation (PVD) is a multi-round challenge-response protocol that produces both an answer and a structured confidence verdict. Two participants, a \textbf{Prover} $\mathcal{P}$ and a \textbf{Verifier} $\mathcal{V}$, each a frozen language model, engage over a multiple-choice question $q$ with candidate answers $\mathcal{A}$.

\textbf{Initialization.} $\mathcal{P}$ selects a candidate answer $\hat{a} \in \mathcal{A}$ and produces a structured proof: a one-sentence \emph{statement} asserting why $\hat{a}$ is correct, a list of 3--5 \emph{atomic sub-claims} (each independently verifiable), and a brief \emph{reasoning} paragraph justifying the chain.

\textbf{Challenge-response loop.} At each round $t \in \{1,\ldots,T\}$, $\mathcal{V}$ examines the current proof transcript and issues one of three verdicts.
\begin{itemize}[leftmargin=1.5em, itemsep=0pt, topsep=0pt]
\item \textsc{Accept}: every sub-claim is specific, verifiable, and mutually consistent. The protocol terminates with acceptance.
\item \textsc{Reject}: a clear logical flaw, factual error, or internal contradiction has been identified. The attempt terminates in rejection.
\item \textsc{Challenge}: $\mathcal{V}$ identifies the single most suspicious or least-justified sub-claim $\tilde{c}$ and poses a targeted question. $\mathcal{P}$ responds by justifying $\tilde{c}$ more rigorously; it may revise $\hat{a}$ if it detects an error. The dialogue proceeds to round $t{+}1$.
\end{itemize}

\textbf{Fatigue.} If $\mathcal{V}$ issues $T$ challenges without a terminal verdict, the attempt ends in a \emph{fatigue rejection}.

\textbf{Retry.} If an attempt ends in Reject or fatigue, the protocol initiates a new attempt (up to $K$ total), providing $\mathcal{P}$ with a brief summary of prior failed attempts as adversarial context.

\textbf{Fallback.} If no attempt among $K$ reaches \textsc{Accept} with no answer revision, the final answer is determined by majority vote over all $K$ attempt final answers.

\textbf{Accept + No Change (ANC).} We distinguish \textbf{Accept + No Change (ANC)}: an attempt is ANC if $\mathcal{V}$ issues \textsc{Accept} and $\mathcal{P}$'s answer was never revised across all challenge rounds. ANC is the primary selective-prediction signal: ANC questions form the reported subset; the remainder are either withheld or answered by majority vote. An \textsc{Accept} with answer revision indicates the prover corrected itself under pressure --- evidence of prior uncertainty --- and is not treated as high-confidence.

\textbf{Coverage and precision.} We evaluate ANC as a binary classifier that predicts ``the answer is correct.'' Treating ANC as the positive prediction, the two metrics are:
\begin{align*}
  \textbf{HC-Cov}  &= \Pr[\textsc{ANC}]                       &&\text{(coverage: fraction of questions reported)} \\
  \textbf{HC-Prec} &= \Pr[\text{correct} \mid \textsc{ANC}]   &&\text{(precision: accuracy on the reported subset)}
\end{align*}
HC-Prec is the standard precision of the ANC classifier; its complement, $\Pr[\text{wrong} \mid \textsc{ANC}]$, is the \emph{selective risk} of \citet{geifman2017selective}. We deliberately report HC-Cov rather than recall ($\Pr[\textsc{ANC} \mid \text{correct}]$) because the non-ANC complement is intended for downstream remediation (escalation to a more capable model, extended compute, or human review) rather than treated as a missed prediction. The \textbf{Gap} between HC-Prec and accuracy on the non-ANC complement quantifies how much information the ANC verdict carries beyond raw accuracy. The same definitions apply to baseline methods by substituting each method's natural high-confidence subset for ANC (full consensus for SC/USC; agent consensus for Debate; stability for Reflexion).

\subsection{Algorithm}

Algorithm~\ref{alg:pvd} gives the pseudocode. Separate histories $h_\mathcal{P}$ and $h_\mathcal{V}$ are maintained for prover and verifier respectively; $\mathcal{V}$ does not observe $\mathcal{P}$'s internal reasoning, only its structured outputs (answer, statement, sub-claims, reasoning paragraph).

\begin{algorithm}[t]
\caption{Prover-Verifier Deliberation (PVD)}
\label{alg:pvd}
\begin{algorithmic}[1]
\Require Question $q$, choices $\mathcal{A}$, prover $\mathcal{P}$, verifier $\mathcal{V}$, fatigue limit $T$, max attempts $K$
\Ensure Final answer $\hat{a}^*$, outcome $\in \{\textsc{ANC},\, \textsc{MajVote}\}$
\State $\mathrm{log} \gets [\,]$
\For{$k = 1,\ldots, K$}
    \State $\mathrm{ctx} \gets \textsc{FailureSummary}(\mathrm{log})$ \Comment{empty on first attempt}
    \State $(\hat{a},\, \mathrm{stmt},\, \mathbf{c},\, \mathrm{rsn}) \gets \mathcal{P}(q,\, \mathcal{A},\, \mathrm{ctx})$ \Comment{initial proof; $\mathbf{c}$ = sub-claims}
    \State $h_\mathcal{P} \gets [\,(q,\hat{a},\mathrm{stmt},\mathbf{c},\mathrm{rsn})\,]$;\quad $h_\mathcal{V} \gets [\,(q,\hat{a},\mathrm{stmt},\mathbf{c},\mathrm{rsn})\,]$
    \State $\Delta \gets 0$ \Comment{answer-change counter}
    \For{$t = 1,\ldots,T$}
        \State $(\mathrm{verd},\, \mathrm{rsn}_\mathcal{V},\, c_t,\, \tilde{c}_t) \gets \mathcal{V}(h_\mathcal{V})$ \Comment{$c_t$: challenge text; $\tilde{c}_t$: targeted sub-claim}
        \State Append $\mathcal{V}$'s response to $h_\mathcal{V}$
        \If{$\mathrm{verd} = \textsc{Accept}$}
            \State $\mathrm{log}.\mathrm{append}((\hat{a},\, \textsc{ANC}\text{ if }\Delta{=}0\text{ else }\textsc{AccChg},\, t))$
            \If{$\Delta = 0$} \textbf{return} $\hat{a},\, \textsc{ANC}$ \EndIf \Comment{high-confidence: exit immediately}
            \State \textbf{break} \Comment{answer changed: log and retry}
        \ElsIf{$\mathrm{verd} = \textsc{Reject}$ \textbf{ or } $t = T$}
            \State $\mathrm{log}.\mathrm{append}((\hat{a},\, \textsc{Reject},\, t))$; \textbf{break}
        \Else \Comment{\textsc{Challenge}}
            \State $(\hat{a}',\, \mathrm{stmt}',\, \mathbf{c}',\, \mathrm{rsn}') \gets \mathcal{P}(h_\mathcal{P},\, \tilde{c}_t,\, c_t)$ \Comment{targeted defence}
            \If{$\hat{a}' \neq \hat{a}$} $\Delta \mathrel{+}= 1$;\quad $\hat{a} \gets \hat{a}'$ \EndIf
            \State Append $\mathcal{P}$'s reply to $h_\mathcal{P}$, $h_\mathcal{V}$
        \EndIf
    \EndFor
\EndFor
\State \textbf{return} $\textsc{MajorityVote}(\mathrm{log}),\, \textsc{MajVote}$
\end{algorithmic}
\end{algorithm}

\textbf{No formal soundness or completeness.} Classical guarantees rest on computational complexity assumptions that do not hold for frozen LLMs. $\mathcal{P}$ may hallucinate or persuasively defend incorrect answers; $\mathcal{V}$ may miss valid challenges or issue spurious rejections. HC-Prec and the \textsc{Accept}/\textsc{Reject} accuracy gap serve as their empirical proxies: HC-Prec measures the effective completeness of the accepted set, while the gap measures how reliably rejection identifies incorrect answers.

\textbf{Separated histories.} $\mathcal{P}$ and $\mathcal{V}$ maintain independent conversation histories; $\mathcal{V}$ observes only $\mathcal{P}$'s structured output (answer, statement, sub-claims, reasoning), not its internal chain-of-thought or prior conversations. This mirrors the zero-knowledge property: the verifier's judgment process is not visible to the prover, reducing the risk that $\mathcal{P}$ optimizes its responses against $\mathcal{V}$'s specific decision boundary. A degenerate single-call variant in which  $\mathcal{P}=\mathcal{V}$ is evaluated as an ablation in §\ref{sec:analysis}.

\textbf{Retry and majority vote.} Classical IP asks a single question: can $\mathcal{P}$ convince $\mathcal{V}$ of $s \in L$? PVD adds a retry mechanism. After each rejection, the protocol initiates a fresh attempt with adversarial context from prior failures, accumulating evidence across up to $K$ trajectories. ANC is declared only if \emph{some} attempt terminates in \textsc{Accept} with no revision; otherwise majority vote aggregates over all $K$ final answers. This trades additional inference calls for robustness to single bad deliberation trajectories, and decouples the high-confidence ANC subset from the majority-vote fallback.

\section{Experiments}
\subsection{Benchmarks and Evaluation Setting}

We evaluate PVD on benchmarks that stress advanced reasoning and expert knowledge. GPQA Diamond is a graduate-level, domain-expert benchmark in biology, physics, and chemistry designed to be difficult for non-experts even with web access \citep{rein2024gpqa}. Humanity's Last Exam is a broad expert-level benchmark intended to probe frontier academic knowledge and calibration \citep{phan2025lastexam}. These settings are appropriate for selective prediction because they expose the distinction between being often correct and knowing when an answer is reliable. In particular, when both prover and verifier operate near their competence ceiling, the collapse of the \textsc{Accept}--\textsc{Reject} accuracy gap is itself informative: it indicates that the deliberation protocol no longer provides a trustworthy selection signal.

\subsection{Baselines and Evaluation Protocol}
\label{sec:methods}

\textbf{Baselines.}
We compare PVD against a one-call reference point and four inference-time reasoning methods that represent the dominant approaches to improving LLM accuracy through additional computation.

\textit{Single-call baselines} establish the one-call compute reference point. Where available, we report a direct chain-of-thought call with no additional protocol. For Sonnet 4.6 we also report a single-call self-deliberation ablation in which one response contains both prover and verifier blocks; unlike a plain direct call, this ablation exposes an ANC signal but does not separate prover and verifier histories.

\textit{Self-Consistency} \citep{wang2023selfconsistency} samples $k{=}8$ independent reasoning chains and returns the majority-vote answer. The agreement rate over the $k$ samples serves as the method's confidence proxy: full consensus (agreement $= 1.0$) identifies questions the model answers consistently, while disagreement flags uncertainty.

\textit{Universal Self-Consistency} \citep{chen2024usc} generates the same $k{=}8$ samples and then prompts an LLM aggregator to select the single most internally consistent candidate, replacing majority vote with LLM-based selection. We use the same $k$ and the same agreement rate as the confidence proxy.

\textit{Multi-agent Debate} \citep{du2024debate} runs three independent model instances that observe each other's answers and revise over two rounds. Whether all agents converge to the same answer (consensus) serves as the natural confidence proxy.

\textit{Reflexion} \citep{shinn2023reflexion} iteratively critiques and refines a single model's answer over up to five trials. A final answer that is both stable across trials and unchanged from the initial response serves as the confidence proxy. Because the original protocol relies on access to a ground-truth oracle to detect failure, and we must avoid that signal at inference time, we substitute a self-consistency recheck (a second independent sample at the same temperature) as the failure detector — the model's two attempts must agree for the trial to be considered stable.

\textbf{Model configurations.}
We evaluate PVD under six prover-verifier configurations, varying model family, relative capability, and protocol parameters. Configurations are chosen to test two design axes. First, \textbf{within-family capability-asymmetric pairing}: prover and verifier from the same model family, with the verifier smaller and more skeptical (Claude Sonnet 4.6 / Haiku; GPT-5.4 / GPT-5.4-mini). Second, \textbf{cross-family pairing}: prover and verifier from different families (Gemini 3.1 Pro / GPT-5.5-pro on GPQA; GPT-5.5 / Gemini 3.1 Pro on HLE), testing whether family-diverse verification improves calibration. A single-call self-deliberation variant — one model acting as both prover and verifier — is evaluated as an ablation to isolate the effect of model separation (§\ref{sec:analysis}).

\textbf{Protocol parameters.}
The fatigue limit $T$ caps the number of verifier turns per attempt; upon reaching $T$ without an \textsc{Accept} or \textsc{Reject} verdict, the attempt terminates as a rejection. The retry limit $K$ controls how many independent attempts are made before falling back to majority vote. Unless otherwise noted, two-model single-attempt PVD configurations use $K{=}1$ and $T{=}12$. The single-call self-deliberation ablation uses one model call with an internal fatigue limit of $T{=}6$. The strongest GPQA cross-family retry configuration uses $K{=}6$ and $T{=}12$ and is examined separately as a cost-performance tradeoff. The prompts used are in \ref{app:prompts}.

\paragraph{Inference cost.}
We measure computational cost in \textbf{LLM calls per question}. Reference points: a single model call (direct chain-of-thought, or the one-call self-deliberation ablation) costs 1 call; Self-Consistency uses $k{=}8$ samples; Debate with 3 agents and 2 rounds uses $\text{agents} \times (\text{rounds}+1) = 9$ calls (one initial statement plus one response per round per agent); Reflexion is bounded by 5 trials and averages $\approx$2.5 calls per question in our runs. A single PVD attempt that terminates on verifier turn $t$ uses $2t$ calls: one initial prover statement, $t$ verifier turns, and $t-1$ challenge-reply pairs between them. The minimum is 2 calls when standard PVD accepts on the first verifier turn; PVD$^\dagger$, which forces a minimum of one challenge before any \textsc{Accept}, costs at least 4 calls. Empirically, single-attempt PVD averages 3--6 calls per question across our configurations; the retry protocol ($K{=}6$) averages $\approx$29. Per-method means are reported in the \textbf{Calls} column of Table~\ref{tab:gpqa-main}. Token-level costs are logged from provider API responses and converted to USD using provider list prices at time of evaluation.

\section{Results}
\label{sec:results}

We report results across four tables. Table~\ref{tab:gpqa-main} compares PVD configurations against all baselines on GPQA Diamond. Table~\ref{tab:gpqa-domain} breaks down the ANC signal by domain. Table~\ref{tab:hle} shows results on Humanity's Last Exam. Table~\ref{tab:verifier} isolates the effect of verifier choice holding the prover fixed.

\subsection{GPQA Diamond}
\label{sec:results-gpqa}

Table~\ref{tab:gpqa-main} reports overall accuracy, the coverage and precision of each method's high-confidence (HC) subset, the resulting gap, and the mean number of LLM calls per question. For PVD, the HC subset is ANC; for baselines, it is the method's natural confidence signal (full consensus for SC/USC; agent consensus for Debate; stability for Reflexion). An example PVD conversation is in \ref{sec:example-conv}.

\begin{table}[h]
\centering
\caption{GPQA Diamond results grouped by prover. \textbf{HC-Cov}: fraction of questions flagged high-confidence. \textbf{HC-Prec}: accuracy on that subset. \textbf{Gap}: HC-Prec minus accuracy on the complement. \textbf{Calls}: mean LLM calls per question. $^*$: SC with extended thinking (Epoch AI benchmark); not directly comparable to PVD runs (standard API). $\dagger$: challenge-first verifier prompt. $\ddagger$: HC coverage $>$90\%, leaving $n{<}20$ in the complement; gap estimate unreliable. }
\label{tab:gpqa-main}
\small
\setlength{\tabcolsep}{3pt}
\begin{tabular}{lllccccc}
\toprule
\textbf{Method} & \textbf{Verifier} & \textbf{HC Signal} & \textbf{Acc} & \textbf{HC-Cov} & \textbf{HC-Prec} & \textbf{Gap} & \textbf{Calls} \\
\midrule
\multicolumn{8}{l}{\textit{Sonnet 4.6 as prover}} \\
Single-call PVD & Self & ANC & 78.8\% & 63\% & 83.9\% & $+13.6$ & 1 \\
SC (k=8) & — & Full consensus & 83.3\% & 72\% & 91.5\% & $+29.0$ & 8 \\
USC (k=8) & Sonnet 4.6 & Full consensus & 81.8\% & 72\% & 91.5\% & $+29.0$ & 9 \\
Debate (3×2) & Sonnet 4.6 & Agent consensus & 83.3\% & 95\% & 85.7\% & $+52.4^{\ddagger}$ & 9 \\
Reflexion & — & Stable, unchanged & 82.3\% & 93\% & 82.6\% & $+4.0^{\ddagger}$ & $\leq$5 \\
PVD & Haiku 4.5 & ANC & 76.8\% & 77\% & 84.2\% & $+32.0$ & $\sim$3 \\
PVD$^\dagger$ & Haiku 4.5 & ANC & 79.8\% & 65\% & 89.9\% & $+29.1$ & $\sim$6 \\
\midrule
\multicolumn{8}{l}{\textit{GPT-5.4 as prover}} \\
Direct & — & — & 72.7\% & \textemdash & \textemdash & \textemdash & 1 \\
SC$^*$ (k=8) & — & Full consensus & 94.9\% & 90\% & 97.2\% & $+23.5^{\ddagger}$ & 8 \\
PVD & GPT-5.4-mini & ANC & 77.8\% & 43\% & 97.6\% & $+34.8$ & $\sim$3 \\
\midrule
\multicolumn{8}{l}{\textit{Gemini 3.1 Pro as prover}} \\
SC$^*$ (k=8) & — & Full consensus & 93.9\% & 94\% & 97.3\% & $+60.9^{\ddagger}$ & 8 \\
PVD & Flash-Lite & ANC & 94.4\% & 57\% & 97.3\% & $+6.6$ & $\sim$4 \\
PVD+retry & GPT-5.5-pro & ANC & 92.9\% & 75\% & 97.3\% & $+17.3$ & $\sim$29 \\
\bottomrule
\end{tabular}
\end{table}

\medskip
\noindent\textbf{A note on overall accuracy.}
The \textbf{Acc} column in Table~\ref{tab:gpqa-main} reports the prover's final answer on \emph{all} questions regardless of verdict. Without a downstream remediation step for non-ANC questions, PVD's full-population accuracy is comparable to the one-call Sonnet self-deliberation ablation (76.8--79.8\% vs.\ 78.8\% for Sonnet~4.6). This is by design: PVD is a \emph{selective prediction} protocol, not an accuracy booster. Its primary output is a structured report/abstain signal. A downstream system can use the Reject or non-ANC verdict as a trigger for escalation---routing the question to a more capable model, extended chain-of-thought compute, or human review before committing to an answer. We do not implement this second stage here, but the ANC verdict directly enables such workflows, and the precision gap ($+29$--$32$pp on the ANC subset vs.\ the complement) quantifies the value of the signal.

\medskip
Table~\ref{tab:gpqa-domain} breaks the ANC signal down by domain for three representative PVD configurations. Biology is the smallest domain ($n{=}19$) and the numbers should be interpreted accordingly.

\begin{table}[h]
\centering
\caption{GPQA Diamond ANC results by domain. \textbf{HC-Cov} $= \Pr[\textsc{ANC}]$: fraction of domain questions assigned ANC. \textbf{HC-Prec} $= \Pr[\text{correct} \mid \textsc{ANC}]$: accuracy on ANC subset. \textbf{Gap}: HC-Prec minus non-ANC accuracy.}
\label{tab:gpqa-domain}
\small
\setlength{\tabcolsep}{5pt}
\begin{tabular}{llccccccccc}
\toprule
 & & \multicolumn{3}{c}{\textbf{Sonnet + Haiku}$^\dagger$} & \multicolumn{3}{c}{\textbf{GPT-5.4 + GPT-5.4-mini}} & \multicolumn{3}{c}{\textbf{Gemini Pro + Flash-Lite}} \\
\cmidrule(lr){3-5}\cmidrule(lr){6-8}\cmidrule(lr){9-11}
\textbf{Domain} & $n$ & HC-Cov & HC-Prec & Gap & HC-Cov & HC-Prec & Gap & HC-Cov & HC-Prec & Gap \\
\midrule
Chemistry & 93 & 49\% & 80\% & $+25.1$ & 16\% & 100\% & $+46.2$ & 45\% & 100\% & $+7.8$ \\
Physics & 86 & 83\% & 96\% & $+15.8$ & 70\% & 98\% & $+9.9$ & 66\% & 100\% & $+6.9$ \\
Biology & 19 & 63\% & 92\% & $+34.5$ & 53\% & 90\% & $+23.3$ & 68\% & 77\% & $+10.3$ \\
\bottomrule
\end{tabular}
\end{table}

\subsection{Humanity's Last Exam}
\label{sec:results-hle}

Table~\ref{tab:hle} reports results for the GPT-5.5 prover / Gemini 3.1 Pro verifier configuration on all 513 HLE questions, broken down by domain. This configuration uses the strongest available models and represents the frontier capability ceiling in our evaluation.

\begin{table}[h]
\centering
\caption{Humanity's Last Exam results (GPT-5.5 prover, Gemini 3.1 Pro verifier, $n{=}513$, $T{=}12$, $K{=}1$). Rows sorted by $n$. \textbf{HC-Cov} $= \Pr[\textsc{ANC}]$; \textbf{HC-Prec} $= \Pr[\text{correct} \mid \textsc{ANC}]$; \textbf{Gap}: HC-Prec minus non-ANC accuracy.}
\label{tab:hle}
\small
\setlength{\tabcolsep}{5pt}
\begin{tabular}{lcccccc}
\toprule
\textbf{Domain} & $n$ & \textbf{Overall Acc} & \textbf{HC-Cov} & \textbf{HC-Prec} & \textbf{Non-ANC Acc} & \textbf{Gap} \\
\midrule
Biology / Medicine & 147 & 41\% & 46\% & 50.0\% & 32.9\% & $+17.1$ \\
Math & 89 & 61\% & 67\% & 76.7\% & 27.6\% & $+49.1$ \\
Humanities / Soc.\ Sci. & 79 & 51\% & 53\% & 61.9\% & 37.8\% & $+24.1$ \\
Computer Science / AI & 66 & 52\% & 53\% & 68.6\% & 32.3\% & $+36.3$ \\
Other & 44 & 32\% & 39\% & 52.9\% & 18.5\% & $+34.4$ \\
Physics & 37 & 27\% & 57\% & 33.3\% & 18.8\% & $+14.6$ \\
Chemistry & 26 & 46\% & 35\% & 77.8\% & 29.4\% & $+48.4$ \\
Engineering & 25 & 40\% & 64\% & 31.2\% & 55.6\% & $-24.3$ \\
\midrule
\textbf{All} & \textbf{513} & \textbf{45.6\%} & \textbf{52\%} & \textbf{59.0\%} & \textbf{31.0\%} & $\mathbf{+27.9}$ \\
\bottomrule
\end{tabular}
\end{table}

\subsection{Verifier Choice and Strictness}
\label{sec:results-verifier}

Table~\ref{tab:verifier} holds the prover fixed and varies the verifier to isolate the contribution of verifier strictness and model pairing. Two provers are shown: Sonnet 4.6 and Gemini 3.1 Pro, each paired with verifiers of increasing independence and strictness.

\begin{table}[h]
\centering
\caption{Effect of verifier choice on ANC calibration, holding prover fixed. \textbf{Avg R}: mean challenge rounds. $\dagger$: challenge-first prompt. \emph{Self}: same model as prover (single-call ablation).}
\label{tab:verifier}
\small
\setlength{\tabcolsep}{5pt}
\begin{tabular}{llccccc}
\toprule
\textbf{Prover} & \textbf{Verifier} & \textbf{HC-Cov} & \textbf{HC-Prec} & \textbf{Non-ANC Acc} & \textbf{Gap} & \textbf{Avg R} \\
\midrule
\multicolumn{7}{l}{\textit{Sonnet 4.6 prover}} \\
Sonnet 4.6 & \emph{Self} & 63\% & 83.9\% & 70.3\% & $+13.6$ & 2.1 \\
Sonnet 4.6 & Haiku & 77\% & 84.2\% & 52.2\% & $+32.0$ & 1.7 \\
Sonnet 4.6 & Haiku$^\dagger$ & 65\% & 89.9\% & 60.9\% & $+29.1$ & 3.0 \\
\midrule
\multicolumn{7}{l}{\textit{Gemini 3.1 Pro prover}} \\
Gemini Pro & Flash-Lite & 57\% & 97.3\% & 90.7\% & $+6.6$ & 1.9 \\
Gemini Pro & GPT-5.5-pro & 75\% & 97.3\% & 80.0\% & $+17.3$ & 14.3 \\
\bottomrule
\end{tabular}
\end{table}

\section{Discussion}
\label{sec:analysis}

\subsection{Argument Defensibility as a Correctness Signal}
\label{sec:disc-real}

The ANC signal is not merely a restatement of the prover's prior confidence. On GPQA Diamond, HC-Prec ranges from 84--98\% across all configurations, consistently above both overall accuracy and the non-ANC complement. But the more telling evidence comes from the structure of errors. 

\textbf{Logical and reasoning errors.} The verifier can identify internal inconsistencies or invalid inferential steps without possessing domain expertise. On GPQA Chemistry, the GPT-5.4-mini verifier achieves HC-Prec 100\% with a $+46.2$pp gap (Table~\ref{tab:gpqa-domain}), despite being a substantially smaller model than the prover. Chemistry questions often require multi-step derivations where a single wrong sign, an incorrect stoichiometric ratio, or a mislabelled reaction intermediate breaks the entire proof chain. A verifier that asks "how does step 2 follow from step 1?" can catch this class of error without knowing the answer itself.

\textbf{Insufficient knowledge.} The second failure mode is harder to catch: the prover selects an answer because it has a confident but wrong prior, rather than a traceable reasoning error. These cases require the verifier to possess independent domain knowledge to probe whether the asserted facts are true. The GPT-5.4-mini verifier's large Chemistry gap but modest Physics gap ($+9.9$pp, Table~\ref{tab:gpqa-domain}) reflects this asymmetry: structural reasoning errors in Chemistry are catchable; subtler physical intuitions in Physics require the verifier to know enough to ask a non-trivial question.

\textbf{Effective Verifier.} The third failure mode is verifier-side: a verifier that issues challenges but accepts any confident-sounding reply is not performing effective verification. The Gemini Flash-Lite verifier illustrates this. Despite achieving HC-Prec 97.3\%, the gap is only $+6.6$pp because non-ANC accuracy is also high (90.7\%): Flash-Lite is rejecting correct answers at roughly the same rate as incorrect ones. 

Together, these three modes predict when PVD should and should not work. If the verifier can identify reasoning errors or has relevant domain knowledge, the ANC gap is large. If the verifier is easily satisfied or lacks the knowledge to sustain adversarial pressure, the gap collapses --- a claim confirmed on HLE (\S\ref{sec:disc-capability}).

\textbf{Does separation matter?} We test a self-deliberation variant (for prompt see \ref{app:self-deliberation-prompt}) in which a single model plays both prover and verifier within a single call. Without requiring a minimum of one challenge the single call always accepts; however, with the additional requirement we find that HC-Prec remains 84\% --- suggesting the structured sub-claim decomposition contributes independently of cross-model independence. Notably, the Accept/Reject gap narrows by $\sim$15pp (from +29pp to +14pp), consistent with a weaker Effective Verifier that cannot challenge claims it generated itself.

\textbf{Defensibility versus stability.}
A related but weaker proxy for confidence is \emph{stability}: does the model give the same answer when re-sampled? Reflexion uses stability across iterations as its high-confidence signal (Table~\ref{tab:gpqa-main}). Yet on GPQA Diamond stability turns out to be only marginally informative: Reflexion flags 93\% of questions as stable, but the precision on that subset (82.6\%) is essentially indistinguishable from the accuracy on the unstable complement. The reason is structural. At standard sampling temperatures, Sonnet 4.6 produces the same multiple-choice letter on most pairs of independent samples, including for many questions it answers incorrectly; consistency in the output channel does not depend on whether the reasoning chain is sound. PVD's ANC verdict requires something strictly stronger: the prover's sub-claim decomposition must survive an adversarial challenge round without revision. The empirical gap between Reflexion's $\sim$+4pp signal and PVD's +29--35pp ANC gap supports our thesis that argument defensibility, not sample agreement, is what carries the load for selective prediction.

\subsection{Verifier Capability and the Effective Verifier Condition}
\label{sec:disc-capability}

Table~\ref{tab:hle-capability} shows how the gap responds to increasing prover and verifier capability. The most striking result in Table~\ref{tab:hle-capability} is not the GPT-5.5 / Gemini 3.1 Pro gap of $+27.9$pp; it is the Sonnet 4.6 / Haiku 4.5 gap of $-7.1$pp --- the ANC questions are \emph{less accurate} than the non-ANC questions. This inversion is the empirical signature of an empty effective region in the sense of \citet{zhang2025human}: Haiku lacks the domain knowledge required to formulate non-trivial challenges on HLE questions, which means that questions where it \emph{can} construct a challenge are precisely the questions where some domain knowledge is available --- but those are not the hard ones. The verifier challenges what it dimly recognizes and accepts what it cannot evaluate at all, inverting the intended selection.

\begin{table}[h]
\centering
\caption{HLE ANC gap by model pairing. \textbf{HC-Cov} $= \Pr[\textsc{ANC}]$; \textbf{HC-Prec} $= \Pr[\text{correct} \mid \textsc{ANC}]$; \textbf{Gap}: HC-Prec minus non-ANC accuracy.}
\label{tab:hle-capability}
\small
\begin{tabular}{llcccc}
\toprule
\textbf{Prover} & \textbf{Verifier} & \textbf{Acc} & \textbf{HC-Cov} & \textbf{HC-Prec} & \textbf{Gap} \\
\midrule
Sonnet 4.6 & Haiku 4.5 & 20.1\% & 31\% & 15.2\% & $-7.1$ \\
Opus 4.6 & Sonnet 4.6 & 40.0\% & 67\% & 41.4\% & $+4.3$ \\
GPT-5.5 & Gemini 3.1 Pro & 45.6\% & 52\% & 59.0\% & $+27.9$ \\
\bottomrule
\end{tabular}
\end{table}

The same pattern appears within GPQA when the verifier is varied with the prover held fixed (Table~\ref{tab:verifier}). Holding Gemini 3.1 Pro as prover, replacing Flash-Lite with GPT-5.5-pro increases average rounds from 1.9 to 14.3 and widens the gap from $+6.6$pp to $+17.3$pp. HC-Prec is identical at 97.3\%; the difference is in the non-ANC accuracy, which drops from 90.7\% to 80.0\%. GPT-5.5-pro sustains multi-round adversarial pressure long enough to reject a more error-enriched complement; Flash-Lite accepts after a single round and provides little additional selection signal.

\subsection{PVD and SC Are Not Measuring the Same Thing}
\label{sec:disc-sc}

\begin{table}[h]
\centering
\caption{Overlap between SC full-consensus and PVD ANC on GPQA Diamond ($n{=}198$). Accuracy columns report correctness within each cell.}
\label{tab:sc-pvd}
\small
\begin{tabular}{lcccc}
\toprule
 & \textbf{$n$} & \textbf{\%} & \textbf{SC Acc} & \textbf{PVD Acc} \\
\midrule
Both HC & 107 & 54\% & 96.3\% & 96.3\% \\
SC only & 35 & 18\% & 77.1\% & 62.9\% \\
PVD only & 22 & 11\% & 72.7\% & 59.1\% \\
Neither & 34 & 17\% & 55.9\% & 58.8\% \\
\bottomrule
\end{tabular}
\end{table}

Self-Consistency and PVD both identify a high-confidence subset from the same base model (Sonnet 4.6), yet their selected subsets overlap only partially (Table~\ref{tab:sc-pvd}). The 54\% overlap at 96.3\% precision confirms that both methods reliably identify the same easy-to-answer questions. The disagreements are more informative. On the 35 questions where SC is confident but PVD rejects, SC accuracy falls to 77.1\% --- well below the 96.3\% joint precision. PVD rejected 8 of the 12 cases (67\%) where SC was confident yet wrong. Conversely, on the 13 questions where PVD accepted but the answer was wrong, SC had already flagged non-consensus on 9 (69\%). Each method catches a distinct class of the other's errors.

The underlying mechanisms are fundamentally different. SC measures \textbf{output consistency}: high agreement means the model reliably generates the same answer across independent samples, which correlates with correctness but can amplify confident hallucinations.  PVD measures \textbf{argument defensibility}: ANC means the prover's sub-claim decomposition survived targeted adversarial challenge without revision, which is sensitive to reasoning structure but blind to consistency across samples. A question where the model always gives the right answer for the wrong reason will pass SC but may fail PVD; a question where the model's samples vary but one reasoning chain is logically airtight may fail SC but pass PVD.

This complementarity has a practical implication. The questions that \emph{both} methods flag as high-confidence achieve 96.3\% precision---the highest subset in Table~\ref{tab:sc-pvd}---while neither method alone reaches this on its respective full coverage. An ensemble signal (ANC $\cap$ full-consensus) trades coverage for a sharper precision operating point, reporting on only $\sim$54\% of questions.

A more aggressive combination---suggested by the structural difference between the signals---would restart PVD until it accepts, analogous to the way SC finds a majority-vote answer across multiple draws. This ``PVD-consistency'' variant would keep re-running the deliberation protocol until the prover earns an ANC verdict, then report. If the protocol is not merely guessing, restarts would succeed more often on genuinely answerable questions and accumulate rejections on genuinely hard ones. The final accepted answer is cherry-picked in the sense that it was selected from multiple attempts, making the nominal ANC verdict optimistic as a calibration estimate; nevertheless, the question of whether structured re-deliberation can recover additional correct answers that single-attempt ANC misses is a natural avenue for future work. Combining PVD with consistency-based sampling is also complementary to remediation workflows, where non-ANC questions trigger more capable models rather than more attempts of the same model.

\section{Limitations}
\label{sec:limitations}

Several limitations bound these conclusions. First, all experiments are conducted on English-language multiple-choice benchmarks with a fixed answer alphabet. This setting provides an unambiguous correctness criterion, but it may not transfer directly to open-ended generation, multilingual tasks, long-form factual synthesis, or settings where multiple answers can be partially correct.

Second, all prover and verifier agents are proprietary API-access models; no open-weight models are evaluated. This choice lets us study current hosted frontier systems, but it limits reproducibility and makes results sensitive to provider-side model updates, decoding defaults, API availability, and undocumented serving changes. The experiments should therefore be interpreted as evaluations of the specified hosted models at the time they were run rather than as claims about all language models.

Third, our selective-prediction analysis uses the protocol's native verdict signals, especially ANC, rather than a learned or distribution-free selector. A full conformal risk-control baseline would require a separate calibration split for GPQA and HLE, which is difficult given the small size of GPQA Diamond and the limited number of complete HLE runs. Similarly, a learned confidence head over features such as ANC, number of challenge rounds, answer changes, and agreement with self-consistency is a natural extension, but would require train/validation splits and would risk overfitting at the sample sizes studied here. We therefore defer conformal and learned-selector comparisons to future work.

Fourth, the baseline comparisons are not exhaustive hyperparameter sweeps. In particular, our Debate baseline uses one fixed configuration of three agents and two rounds; sweeping the number of agents, number of rounds, or consensus thresholds would require substantial additional API calls. The same caveat applies more generally to operating-point selection for baseline confidence signals: we compare against natural confidence proxies rather than tuning every method to every possible coverage level.

Fifth, we measure computational cost in LLM calls and token cost rather than wall-clock latency. Wall-clock latency depends strongly on provider queueing, server load, rate limits, batching, and parallelization choices, making it difficult to compare fairly across APIs. Call count and token usage are more stable proxies for inference-time compute, but they do not capture all deployment-relevant costs.

Finally, PVD assumes an honest prover that defends its genuine best answer rather than strategically optimizing against the verifier. It is not an adversarial robustness guarantee, an AI-control protocol, or a formal interactive-proof system. A strategic prover, a colluding model pair, or a verifier operating outside its domain competence can defeat, collapse, or invert the ANC signal. This failure mode is visible in the HLE results and is central to our interpretation: PVD is useful only when the verifier remains inside an effective region where targeted challenges are informative.

\section{Conclusion}
\label{sec:conclusion}

We introduced Prover-Verifier Deliberation (PVD), an inference-time protocol that instantiates the challenge-response structure of interactive proofs with frozen LLMs to produce a selective-prediction signal. The core mechanism is simple: a prover decomposes a candidate answer into atomic sub-claims and defends them under adversarial challenge; a verifier terminates with Accept, Challenge, or Reject. Questions where the prover's answer is accepted without revision --- Accept + No Change (ANC) --- form a high-confidence reported subset. We characterize this subset empirically rather than theoretically, asking not whether formal soundness and completeness hold, but whether the verifier's verdict separates reliable answers from unreliable ones.

Three findings emerge from systematic evaluation on GPQA Diamond and Humanity's Last Exam. First, the ANC signal is genuinely predictive. On GPQA Diamond, HC-Prec ranges from 84--98\% across five model configurations, with gaps over the complement of $+6.6$ to $+34.8$pp. The signal is not a restatement of the prover's prior confidence: SC and PVD  are measuring structurally different things. Second, verifier strictness and domain competence are the primary drivers of calibration quality. A permissive verifier that accepts after one or two rounds provides little selection signal even at high HC-Prec; a verifier that sustains multi-round adversarial pressure forces incorrect provers to contradict themselves. Third, the protocol provides a deployment diagnostic. On HLE, where Sonnet 4.6 paired with Haiku 4.5 produces an \emph{inverted} ANC gap, the verifier's lack of domain knowledge causes it to accept exactly the questions it cannot evaluate. 

The broader implication is architectural. Selective prediction from LLMs does not require access to ground-truth labels, trained reward models, or ensemble budgets comparable to SC. A two-model deliberation at roughly three to six calls per question achieves calibration comparable to eight-sample self-consistency at a fraction of the cost, with a qualitatively different error profile. More importantly, the protocol knows when it cannot be trusted: the ANC gap collapse on HLE is a ground-truth-free signal that the verifier has reached its competence ceiling.

\bibliographystyle{abbrvnat}
\bibliography{ref}


\appendix

\section{Technical appendices and supplementary material}

\subsection{Further Related Work}
\subsubsection{Selective Prediction, Abstention, and Calibration}
\label{sec:selective-prediction}

Selective prediction studies systems that may abstain on uncertain inputs in order to improve accuracy on the accepted subset. The classical reject-option view formalizes a tradeoff between error and rejection rate \citep{chow1970Optimum}. Modern selective classification casts this as a
risk--coverage problem: a model should minimize error on the covered region while abstaining elsewhere \citep{geifman2017selective,pugnana2024deep}. Subsequent work trains selective neural networks end-to-end with an integrated reject option \citep{geifman2019selectivenet}, and extends selective prediction to question answering under distribution shift \citep{kamath2020selective}. This literature provides the evaluation lens for our work: the relevant object is not only full-coverage accuracy, but the precision achieved at a given coverage level.

Calibration asks whether a model's confidence corresponds to empirical correctness. Neural networks are often miscalibrated \citep{guo2017calibration}, and generative language models introduce additional calibration difficulties because probability mass is distributed over many possible completions \citep{jiang2021know}. In the LLM setting, several papers ask whether models ``know what they know'' \citep{kadavath2022language, lin2022teaching, tian2023calibration, xiong2024express}.
Other approaches estimate reliability from disagreement among generations. SelfCheckGPT detects hallucination by checking whether sampled responses remain factually consistent \citep{manakul2023selfcheckgpt}. 

These methods are close in spirit to PVD because they treat inconsistency as evidence of unreliability. The distinction is that PVD does not rely only on independent samples or scalar confidence. Instead, the confidence signal is procedural: a prover decomposes and defends an answer, while a verifier searches for weaknesses and issues an explicit \textsc{Accept}, \textsc{Challenge}, or \textsc{Reject} verdict. Thus, the selector is not a thresholded probability, but the outcome of a structured verification dialogue.

\subsubsection{Inference-Time Reasoning and Self-Aggregation}
\label{sec:inference-time-reasoning}
A large body of work improves LLM performance by spending additional computation at inference time. Chain-of-thought prompting elicits intermediate reasoning steps and improves accuracy on multi-step reasoning tasks \citep{wei2022chain}. Self-Consistency samples multiple independent reasoning paths and aggregates by majority vote \citep{wang2023selfconsistency}. Universal Self-Consistency extends this idea to free-form outputs by using an LLM to select the most consistent answer among sampled candidates \citep{chen2024usc}. Reflexion and Self-Refine use verbal feedback loops to critique and revise model outputs \citep{shinn2023reflexion,madaan2023selfrefine}, while Tree of Thoughts structures inference as search over possible reasoning states \citep{yao2023tot}.

Verification-oriented inference-time methods are also relevant. Chain-of-Verification asks models to generate and answer verification questions about their own outputs \citep{dhuliawala2024chain}, and self-verification methods ask models to check or re-derive candidate answers before selecting a final output \citep{weng2023selfverification}. Recent theoretical work studies sampling-based test-time scaling through the lens of confidence estimation, relating self-consistency and internal probability estimates and proposing hybrid approaches that reduce sampling cost \citep{zhou2025bridging}.

PVD differs from these methods in objective. Most inference-time reasoning methods optimize full-coverage answer accuracy; any reliability signal is a byproduct of agreement, convergence, or successful refinement. PVD instead makes the reliability signal an explicit output of the protocol. The central empirical question is whether the verifier's verdict defines a high-precision accepted subset at useful coverage and cost. This makes PVD closer to selective prediction than to ordinary test-time ensembling: the protocol is evaluated not only by the answer it returns, but also by whether it knows when not to report.

\subsubsection{Verification, Critique, Judges, and Debate}
\label{sec:verification}
A separate literature uses verifiers or critics to evaluate generated reasoning. \citet{cobbe2021training} train verifiers to score candidate solutions to math word problems, showing that learned verification can improve sample efficiency. \citet{lightman2023letsverify} extend this idea to process supervision, where intermediate reasoning steps are scored rather than only final answers. \citet{saunders2022selfcritique} study self-critique as a way to help humans identify flaws in model outputs. 

PVD also uses a verifier, but not as a static reward model, reranker, or post-hoc judge. The verifier actively participates in the dialogue: it examines the prover's decomposition, challenges weak or unsupported subclaims, and terminates with a structured verdict. This distinction matters because the verdict is not merely an evaluation score; it is the mechanism by which the
system chooses between reporting and abstaining.

Multi-agent debate is another closely related direction. Debate methods cast inference as an exchange among multiple agents, often with answers revised after peer arguments \citep{du2024debate}. Some work finds that debate can help weaker or non-expert judges adjudicate difficult questions \citep{khan2024debate,michael2023debate}, while other work finds that short debate protocols do not reliably improve judge accuracy on hard reading comprehension tasks \citep{parrish2022two}. Recent work also studies how to encourage diverse perspectives and avoid premature convergence in multi-agent settings \citep{liang2024encouraging}.

Our protocol is deliberately asymmetric. In debate, agents usually argue toward a shared answer or persuade a judge among competing claims. In PVD, the prover commits to a candidate answer and the verifier's role is exclusively skeptical: to identify whether the argument is sufficient, incomplete, or wrong. This asymmetry sacrifices some collaborative refinement, but yields a cleaner selective-prediction signal: accepted unchanged answers are treated as a high-confidence subset, while challenged or rejected answers naturally induce abstention.

\subsection{Implementation Details}

\subsubsection{Baselines}
For the primary Sonnet GPQA comparison, baselines use Claude Sonnet 4.6 to control for model capability. Self-Consistency and USC use $k{=}8$; Debate uses 3 agents and 2 rounds; Reflexion uses a maximum of 5 trials. Additional GPT-5.4 and Gemini 3.1 Pro rows are included as robustness comparisons where matching external or direct baselines are available.

\subsubsection{Statistical Significance}
\label{sec:stat-sig}
\begin{table}[h]
\centering
\caption{Clean RTTR runs on GPQA Diamond ($n{=}198$) with 95\% confidence intervals and a gap significance test. \textbf{Acc}, \textbf{HC-Cov}, and \textbf{HC-Prec} use Wilson score intervals; \textbf{Gap} (HC-Prec minus non-HC accuracy) uses a percentile bootstrap (10{,}000 resamples). \textbf{Gap $p$} is a two-sided Fisher's exact test of $H_0$: HC and non-HC subsets are equally accurate. A Gap CI excluding 0 (equivalently $p<0.05$) indicates the high-confidence subset is significantly more accurate than its complement. $^*$: SC with extended thinking (Epoch AI benchmark).}
\label{tab:rttr-stats}
\small
\setlength{\tabcolsep}{4pt}
\begin{tabular}{lccccc}
\toprule
\textbf{Run} & \textbf{Acc} & \textbf{HC-Cov} & \textbf{HC-Prec} & \textbf{Gap (95\% CI)} & \textbf{Gap $p$} \\
\midrule
PVD (standard) & 74.2 [67.7, 79.8] & 54.5 [47.6, 61.3] & 93.5 [87.2, 96.8] & +42.4 [+31.0, +53.5] & $<$0.001 \\
PVD (min-1) & 76.3 [69.9, 81.7] & 51.0 [44.1, 57.9] & 91.1 [83.9, 95.2] & +30.3 [+18.9, +41.5] & $<$0.001 \\
PVD (self-play) & 80.3 [74.2, 85.2] & 63.1 [56.2, 69.5] & 94.4 [88.9, 97.3] & +38.2 [+26.0, +50.4] & $<$0.001 \\
PVD+retry & 79.8 [73.7, 84.8] & 79.8 [73.7, 84.8] & 88.6 [82.7, 92.7] & +43.6 [+27.2, +60.0] & $<$0.001 \\
Debate (3$\times$2) & 83.3 [77.5, 87.9] & 95.5 [91.6, 97.6] & 85.7 [80.0, 90.0] & +52.4 [+17.4, +85.6] & $<$0.001 \\
Reflexion & 82.3 [76.4, 87.0] & 92.9 [88.5, 95.7] & 82.6 [76.5, 87.4] & +4.0 [-17.7, +29.1] & 0.717 \\
SC ($k{=}8$)$^*$ & 87.4 [82.0, 91.3] & 81.3 [75.3, 86.1] & 96.3 [92.1, 98.3] & +47.6 [+30.9, +64.1] & $<$0.001 \\
USC ($k{=}8$) & 85.4 [79.8, 89.6] & 81.3 [75.3, 86.1] & 96.3 [92.1, 98.3] & +58.4 [+42.2, +74.1] & $<$0.001 \\
Single-call PVD & 75.8 [69.3, 81.2] & 59.6 [52.6, 66.2] & 83.1 [75.3, 88.8] & +18.1 [+5.7, +30.7] & 0.004 \\
\bottomrule
\end{tabular}
\end{table}

Table~\ref{tab:rttr-stats} reports the clean RTTR runs on GPQA Diamond ($n{=}198$) with 95\% confidence intervals and a formal test of the selection gap. Accuracy, HC-Cov, and HC-Prec use Wilson score intervals; the Gap (HC-Prec minus non-HC accuracy) uses a percentile bootstrap over $10{,}000$ resamples. We additionally test each gap with a two-sided Fisher's exact test of $H_0$: the high-confidence and complement subsets are equally accurate, on the $2\times2$ table of (HC vs.\ non-HC) $\times$ (correct vs.\ incorrect). Fisher's exact test is used rather than a normal-approximation $z$-test because some complements are small (e.g.\ Debate leaves $n{<}10$ in the non-HC set). The two procedures agree: every PVD configuration has a Gap CI bounded well above zero and $p<0.001$, while Reflexion's CI brackets zero and its gap is not significant ($p{=}0.72$) --- consistent with stability being a far weaker selective-prediction signal than argument defensibility. Pairwise McNemar tests across methods (paired on the same $198$ questions) are available through \texttt{python -m rttr.stats -{}-pairs}.

\subsection{Prompts}
\label{sec:prompts}

All models receive a system prompt that defines their role and the required JSON response format. The prover and verifier prompts are given below verbatim.

\subsubsection*{Prover System Prompt}

\begin{quote}
\small\ttfamily
You are a Prover (P) in an Interactive Zero-Knowledge Proof Protocol solving a multiple-choice science question.

Your role:\\
- Select the single best answer (A, B, C, or D)\\
- Break your reasoning into discrete, verifiable sub-claims\\
- Each sub-claim must be atomic and independently checkable\\
- When challenged on a specific sub-claim, provide deeper justification for THAT claim only\\
- Be consistent --- never contradict your previous statements\\
- If you made an error, correct it explicitly and revise your answer letter

Respond in this exact JSON format:\\
\{\\
\hspace*{1em}"answer": "A" | "B" | "C" | "D",\\
\hspace*{1em}"statement": "one sentence explaining why you chose this answer",\\
\hspace*{1em}"subclaims": ["atomic claim 1", "atomic claim 2", ...],\\
\hspace*{1em}"reasoning": "step-by-step explanation supporting your answer and sub-claims"\\
\}
\end{quote}

\subsubsection*{Standard Verifier System Prompt}
\label{app:prompts}
\begin{quote}
\small\ttfamily
You are a Verifier (V) in an Interactive Zero-Knowledge Proof Protocol evaluating a multiple-choice science answer.

Your role:\\
- Evaluate the Prover's reasoning: Accept, Reject, or Challenge\\
- ACCEPT if the reasoning is logically sound and internally consistent\\
- REJECT if you find a logical flaw, factual error, or internal contradiction\\
- CHALLENGE if one specific sub-claim needs deeper justification\\
- When challenging, target the SINGLE most suspicious or unclear sub-claim\\
- Do not challenge things that are obviously correct\\
- Be rigorous but efficient --- unnecessary challenges waste the fatigue budget

You are evaluating reasoning quality, not checking against a known answer.

Respond in this exact JSON format:\\
\{\\
\hspace*{1em}"verdict": "Accept" | "Reject" | "Challenge",\\
\hspace*{1em}"reasoning": "why you made this decision",\\
\hspace*{1em}"challenge": "your specific question (only if verdict is Challenge, else null)",\\
\hspace*{1em}"challenged\_claim": "the exact sub-claim being challenged (only if Challenge, else null)"\\
\}
\end{quote}

\subsubsection*{Challenge-First Verifier System Prompt ($\dagger$)}

The challenge-first variant (PVD$^\dagger$) uses a stricter verifier that defaults to challenging rather than accepting:

\begin{quote}
\small\ttfamily
You are a Verifier (V) in an Interactive Zero-Knowledge Proof Protocol evaluating expert-level questions that are extremely difficult --- harder than PhD qualifying exams.

Your role:\\
- Evaluate the Prover's reasoning: Accept, Reject, or Challenge\\
- Default to CHALLENGE: these questions are hard enough that almost every answer has at least one sub-claim worth probing\\
- ACCEPT only if every sub-claim is specific, verifiable, and you are highly confident the reasoning is airtight --- vague or hand-wavy justifications do not merit acceptance\\
- REJECT if you find a clear logical flaw, factual error, or internal contradiction\\
- CHALLENGE whenever any sub-claim is asserted without sufficient justification, relies on memorized facts you cannot verify, skips logical steps, or uses imprecise language that could hide an error\\
- Target the SINGLE most suspicious, weakest, or least-justified sub-claim\\
- Never accept reasoning that merely sounds plausible --- demand rigor

[JSON format identical to standard verifier.]
\end{quote}

\subsubsection*{Self-Deliberation System Prompt (ablation)}
\label{app:self-deliberation-prompt}

The self-deliberation variant uses a \emph{single} model call that plays both Prover and Verifier roles in sequence, using delimited blocks to structure internal deliberation:

\begin{quote}
\small\ttfamily
You are simulating an Interactive Zero-Knowledge Proof (ZKP) deliberation on a multiple-choice science question. You will play both roles in sequence within this single response:

\textbf{[PROVER]} --- Selects the best answer and breaks reasoning into atomic, independently verifiable sub-claims.\\
\textbf{[VERIFIER]} --- Critically evaluates the prover's reasoning without reference to any external answer key. Decides: Accept, Reject, or Challenge.

Protocol rules:\\
1. Generate one [PROVER] block, then one [VERIFIER] block. That is one round.\\
2. The verifier MUST Challenge in round 1 --- pick the single weakest or least justified sub-claim and demand rigorous justification. Do not Accept or Reject on the first round.\\
3. From round 2 onwards: if the verifier Challenges, generate the next [PROVER] block (responding to that specific challenge only), then another [VERIFIER] block.\\
4. Stop when the verifier issues Accept or Reject, or after \{fatigue\} rounds (forced Reject --- output FINAL\_VERDICT: Reject(fatigue)).\\
5. The verifier must be genuinely adversarial --- default to Challenge. Accept only when every sub-claim is specific, the logic is airtight, and there is nothing left worth probing. Reject if there is a clear logical flaw or factual error.\\
6. The prover may revise its answer if a challenge exposes a mistake.

Use this exact format for each block:

[PROVER]\\
\{\\
\hspace*{1em}"answer": "A" | "B" | "C" | "D",\\
\hspace*{1em}"statement": "one sentence explaining your choice",\\
\hspace*{1em}"subclaims": ["atomic claim 1", "atomic claim 2", ...],\\
\hspace*{1em}"reasoning": "step-by-step explanation"\\
\}\\
{[}/PROVER{]}

[VERIFIER]\\
\{\\
\hspace*{1em}"verdict": "Accept" | "Reject" | "Challenge",\\
\hspace*{1em}"reasoning": "why you made this decision",\\
\hspace*{1em}"challenge": "your specific question targeting one sub-claim (null if not Challenge)",\\
\hspace*{1em}"challenged\_claim": "the exact sub-claim being challenged (null if not Challenge)"\\
\}\\
{[}/VERIFIER{]}

After the final [VERIFIER] block, output exactly:\\
FINAL\_ANSWER: \textlangle letter\textrangle\\
FINAL\_VERDICT: \textlangle Accept|Reject|Reject(fatigue)\textrangle
\end{quote}

\subsection{Example Conversation}
\label{sec:example-conv}

The following illustrates a complete 2-round PVD interaction on a GPQA Diamond question from the Molecular Biology domain (Sonnet~4.6 prover, Haiku~4.5 verifier, challenge-first prompt, correct answer D).

\medskip
\noindent\textbf{Question.}
\textit{There has been an outbreak of a viral infectious disease in your city. You have been asked to design a molecular diagnostic kit for quick detection of this retrovirus. How would you go about designing this kit?}

\smallskip
\noindent(A) Identify the virus by DNA sequencing; develop a PCR kit.\\
(B) Identify the virus from patient symptoms; design a nested PCR kit.\\
(C) Identify IgG antibodies; develop an ELISA kit targeting IgG antibodies.\\
(D) Identify the virus by cDNA sequencing; develop a real-time PCR kit.

\bigskip
\noindent\textbf{Prover — Round 1} \hfill\textit{answer: D}

\smallskip
\noindent\textit{Statement:} Since the pathogen is a retrovirus (RNA-based), you must first convert its RNA genome to cDNA via reverse transcription before sequencing, and then develop a real-time RT-PCR kit for rapid, quantitative, and accurate molecular diagnosis.

\smallskip
\noindent\textit{Sub-claims:}
\begin{enumerate}[leftmargin=2em,itemsep=1pt,topsep=2pt]
  \item Retroviruses have an RNA genome, not a DNA genome.
  \item To sequence a retroviral genome, RNA must first be reverse-transcribed into cDNA.
  \item Real-time PCR (RT-qPCR) is the gold-standard molecular diagnostic method for RNA viruses.
  \item Option A is incorrect: direct DNA sequencing is inappropriate for an RNA retrovirus without reverse transcription.
  \item Option C is incorrect: IgG ELISA detects antibody response (serological), not the virus itself.
\end{enumerate}

\bigskip
\noindent\textbf{Verifier — Round 1} \hfill\textit{verdict: \textsc{Challenge}}

\smallskip
\noindent\textit{Reasoning:} The fundamentals are sound, but sub-claim~2 requires further justification. Modern sequencing platforms (Nanopore, Illumina) can directly sequence RNA without reverse transcription. The prover has not explained why cDNA sequencing specifically is mandatory as the first step, rather than direct RNA sequencing or metagenomics.

\smallskip
\noindent\textit{Challenge (targeting sub-claim 2):} You state that ``RNA must first be reverse-transcribed into cDNA before sequencing.'' However, modern platforms can sequence RNA directly. Why is cDNA sequencing specifically mandatory for diagnostic kit design rather than direct RNA sequencing? And does Option D's phrase ``real-time PCR kit'' unambiguously imply RT-PCR?

\bigskip
\noindent\textbf{Prover — Round 2} \hfill\textit{answer: D (unchanged)}

\smallskip
\noindent\textit{Sub-claims:}
\begin{enumerate}[leftmargin=2em,itemsep=1pt,topsep=2pt]
  \item Direct RNA sequencing exists but is not the standard validated workflow for clinical outbreak diagnostic kit design under current regulatory norms.
  \item cDNA sequencing via reverse transcription is the established, reproducible, and validated method for characterizing RNA virus genomes in diagnostic contexts.
  \item ``Real-time PCR'' in clinical virology is understood to mean real-time RT-PCR, as standard qPCR cannot amplify RNA templates.
  \item Option D is the only choice that correctly combines RNA-to-cDNA conversion with a sensitive, quantitative, closed-tube molecular detection method.
\end{enumerate}

\bigskip
\noindent\textbf{Verifier — Round 2} \hfill\textit{verdict: \textsc{Accept}}

\smallskip
\noindent\textit{Reasoning:} The prover now correctly distinguishes between what is technically possible (direct RNA sequencing exists) and what is standard validated clinical practice (cDNA sequencing followed by RT-PCR). The response appropriately anchors the answer to regulatory and epidemiological norms rather than claiming absolute necessity. The clarification that ``real-time PCR'' is a clinical convention for RT-qPCR is correct and addresses the ambiguity. The comparative evaluation of options is sound. \textbf{Result: ANC} (answer D accepted with no revision).


\newpage
\section*{NeurIPS Paper Checklist}

\begin{enumerate}

    \item {\bf Claims}
    \item[] Question: Do the main claims made in the abstract and introduction accurately reflect the paper's contributions and scope?
    \item[] Answer: \answerYes{} 
    \item[] Justification: The abstract and introduction frame PVD as an empirical selective-prediction protocol, and the results tables report the GPQA Diamond main experiment, robustness experiments, and HLE failure modes within that scope.

    \item {\bf Limitations}
    \item[] Question: Does the paper discuss the limitations of the work performed by the authors?
    \item[] Answer: \answerYes{} 
    \item[] Justification: Section~\ref{sec:limitations} discusses the English-only multiple-choice setting, proprietary API models, absence of open-weight model experiments, lack of conformal and learned-selector baselines, limited debate hyperparameter exploration, cost measurement choices, and the honest-prover assumption.

\item {\bf Theory assumptions and proofs}
    \item[] Question: For each theoretical result, does the paper provide the full set of assumptions and a complete (and correct) proof?
    \item[] Answer: \answerNA{} 
    \item[] Justification: The paper uses interactive proof theory as conceptual motivation and explicitly states that formal soundness and completeness guarantees do not transfer to frozen LLMs; it does not claim a new theorem or proof.

    \item {\bf Experimental result reproducibility}
    \item[] Question: Does the paper fully disclose all the information needed to reproduce the main experimental results of the paper to the extent that it affects the main claims and/or conclusions of the paper (regardless of whether the code and data are provided or not)?
    \item[] Answer: \answerYes{} 
    \item[] Justification: Sections~\ref{sec:protocol}--\ref{sec:results} specify the protocol, benchmarks, model pairings, fatigue and retry parameters, confidence signals, and cost metric; code and result logs are provided as supplemental artifacts.

\item {\bf Open access to data and code}
    \item[] Question: Does the paper provide open access to the data and code, with sufficient instructions to faithfully reproduce the main experimental results, as described in supplemental material?
    \item[] Answer: \answerYes{} 
    \item[] Justification: The implementation scripts, prompts, and result files are included in the supplemental material. GPQA Diamond and HLE are existing benchmark datasets and must be accessed under their original distribution terms.

\item {\bf Experimental setting/details}
    \item[] Question: Does the paper specify all the training and test details (e.g., data splits, hyperparameters, how they were chosen, type of optimizer) necessary to understand the results?
    \item[] Answer: \answerYes{} 
    \item[] Justification: The experiments use frozen API models rather than training, so optimizer details are not applicable. Sections~\ref{sec:methods}--\ref{sec:results} describe benchmarks, model configurations, fatigue/retry parameters, baseline settings, and high-confidence criteria.

\item {\bf Experiment statistical significance}
    \item[] Question: Does the paper report error bars suitably and correctly defined or other appropriate information about the statistical significance of the experiments?
    \item[] Answer: \answerYes{} 
    \item[] Justification: Appendix~\ref{sec:stat-sig} reports 95\% confidence intervals (Wilson for accuracy, coverage, and precision; percentile bootstrap for the selection gap) and a two-sided Fisher's exact test that the gap is non-zero; pairwise McNemar tests across methods are available in the released code. The main tables additionally report exact counts and flag small complements where gap estimates are unreliable.
\item {\bf Experiments compute resources}
    \item[] Question: For each experiment, does the paper provide sufficient information on the computer resources (type of compute workers, memory, time of execution) needed to reproduce the experiments?
    \item[] Answer: \answerNo{} 
    \item[] Justification: All model inference used hosted APIs, so provider-side hardware and wall-clock latency are not meaningful to report (latency depends on queueing, rate limits, and server load). In their place, the paper reports the deployment-relevant compute proxies: LLM calls per question and estimated token cost.
    
\item {\bf Code of ethics}
    \item[] Question: Does the research conducted in the paper conform, in every respect, with the NeurIPS Code of Ethics \url{https://neurips.cc/public/EthicsGuidelines}?
    \item[] Answer: \answerYes{} 
    \item[] Justification: The work evaluates hosted LLMs on public benchmark questions and does not involve human subjects, private data, model release, or collection of sensitive information.

\item {\bf Broader impacts}
    \item[] Question: Does the paper discuss both potential positive societal impacts and negative societal impacts of the work performed?
    \item[] Answer: \answerYes{} 
    \item[] Justification: The paper discusses the intended benefit of selective prediction and abstention, and Section~\ref{sec:limitations} discusses the risk of overtrust when the verifier is outside its competence or when a strategic prover defeats the protocol.
    
\item {\bf Safeguards}
    \item[] Question: Does the paper describe safeguards that have been put in place for responsible release of data or models that have a high risk for misuse (e.g., pre-trained language models, image generators, or scraped datasets)?
    \item[] Answer: \answerNA{} 
    \item[] Justification: The paper does not release a new dataset, pre-trained model, or scraped corpus; it releases evaluation code and aggregate/results artifacts for existing benchmarks.

\item {\bf Licenses for existing assets}
    \item[] Question: Are the creators or original owners of assets (e.g., code, data, models), used in the paper, properly credited and are the license and terms of use explicitly mentioned and properly respected?
    \item[] Answer: \answerNo{} 
    \item[] Justification: The paper cites GPQA Diamond and HLE and names the proprietary API models used; however, the main text does not enumerate all dataset licenses or API terms of service. No open-weight models were used.
\item {\bf New assets}
    \item[] Question: Are new assets introduced in the paper well documented and is the documentation provided alongside the assets?
    \item[] Answer: \answerYes{} 
    \item[] Justification: The reproduction code, prompts, evaluation package, and the Claude Code PVD skill are released in a public, MIT-licensed GitHub repository (\url{https://github.com/jsedoc/prover-verifier-deliberation}) with a README documenting usage.

\item {\bf Crowdsourcing and research with human subjects}
    \item[] Question: For crowdsourcing experiments and research with human subjects, does the paper include the full text of instructions given to participants and screenshots, if applicable, as well as details about compensation (if any)? 
    \item[] Answer: \answerNA{} 
    \item[] Justification: The paper does not involve crowdsourcing experiments or human subjects.

\item {\bf Institutional review board (IRB) approvals or equivalent for research with human subjects}
    \item[] Question: Does the paper describe potential risks incurred by study participants, whether such risks were disclosed to the subjects, and whether Institutional Review Board (IRB) approvals (or an equivalent approval/review based on the requirements of your country or institution) were obtained?
    \item[] Answer: \answerNA{} 
    \item[] Justification: The work does not involve human subjects, so IRB review is not applicable.

\item {\bf Declaration of LLM usage}
    \item[] Question: Does the paper describe the usage of LLMs if it is an important, original, or non-standard component of the core methods in this research? Note that if the LLM is used only for writing, editing, or formatting purposes and does \emph{not} impact the core methodology, scientific rigor, or originality of the research, declaration is not required.
    \item[] Answer: \answerYes{} 
    \item[] Justification: The use of LLMs is the core experimental object: Sections~\ref{sec:protocol}--\ref{sec:methods} describe frozen LLMs acting as provers, verifiers, and baseline agents.

\end{enumerate}

\end{document}